\documentclass{article}
\pdfoutput=1

\usepackage[preprint]{neurips_2019}

\usepackage[utf8]{inputenc} 
\usepackage[T1]{fontenc}    
\usepackage{hyperref}       
\usepackage{url}            
\usepackage{booktabs}       
\usepackage{amsfonts}       
\usepackage{nicefrac}       
\usepackage{microtype}      
\usepackage{graphicx}
\usepackage{amsmath}
\usepackage{hyperref}
\usepackage[symbol]{footmisc}
\usepackage{blindtext}
\usepackage{enumitem}
\usepackage{setspace}

\hypersetup{
  colorlinks   = true, 
  urlcolor     = blue, 
  linkcolor    = blue, 
  citecolor   = red 
}

\title{NEURO-DRAM: a 3D recurrent visual attention model for interpretable neuroimaging classification}

\author{%
  David A.~Wood\\
  King's College London\\
  \texttt{david.wood@kcl.ac.uk} \\
   \And
   James H. Cole\thanks{Joint senior authors} \\
   King's College London \\
   University College London \\
   \texttt{james.cole@ucl.ac.uk} \\
   \And
   Thomas C. Booth \footnotemark[1] \\
   King's College London\\
   King's College Hospital \\
   \texttt{thomas.booth@kcl.ac.uk} \\
}

\begin{document}

\maketitle
\vspace{-2em}
\setcounter{footnote}{0}
\begin{center}
    For the Alzheimer's Disease Neuroimaging Initiative
\end{center}
\begin{abstract}
 Deep learning is attracting significant interest in the neuroimaging community as a means to diagnose psychiatric and neurological disorders from structural magnetic resonance images. However, there is a tendency amongst researchers to adopt architectures optimized for traditional computer vision tasks, rather than design networks customized for neuroimaging data. We address this by introducing NEURO-DRAM, a 3D recurrent visual attention model tailored for neuroimaging classification. The model comprises an agent which, trained by reinforcement learning, learns to navigate through volumetric images, selectively attending to the most informative regions for a given task. When applied to Alzheimer's disease prediction, NEURO-DRAM achieves state-of-the-art classification accuracy on an out-of-sample dataset, significantly outperforming a baseline convolutional neural network. When further applied to the task of predicting which patients with mild cognitive impairment will be diagnosed with Alzheimer's disease within two years, the model achieves state-of-the-art accuracy with no additional training. Encouragingly, the agent learns, without explicit instruction, a search policy visiting standardized radiological hallmarks of Alzheimer’s disease, suggesting a route to automated biomarker discovery for more poorly understood disorders.
 
\end{abstract}

\section{Introduction}

Neuroimaging has long held promise for automating diagnostic and prognostic decisions for psychiatric and neurological disorders, based on statistical analysis of brain structure and function. However, the clinical adoption of automated neuroimaging analysis has been limited to date, in part because finding subtle and diffuse patterns in high-dimensional brain images that are clinically-meaningful is challenging, and in part because an intuitive understanding of how automated decisions are reached has been lacking. In other words, the performance levels and interpretability of neuroimaging analysis have generally not reached clinically-acceptable thresholds.

In recent years, deep-learning based approaches have attracted substantial interest as a way to improve model performance and provide new approaches to interpreting modelling decisions. For example, a number of papers have demonstrated accurate classification of disorders such as Alzheimer's disease from structural MRI scans using convolutional neural networks (Spasov et al. (2019), Wegmayr et al. (2017), Korolev et al. (2017), Cheng et al. (2017), B\"{o}hle et al. (2019)). Despite this growing interest, only modest improvements in model performance have been achieved since the first demonstration (Vieira et al. (2017)), with limited architectural variety appearing in the literature. We ascribe this to a trend within the neuroimaging community to adopt those architectures that have proved successful in computer vision competitions such as the ImageNet Large-Scale Visual Recognition Challenge, without considering the peculiarities of neuroimaging data. However, this repurposing of architectures is potentially problematic for several reasons:

i) The larger number of parameters inherent to 3D networks increases the likelihood of over-fitting to training data. This is exacerbated by the relative paucity of labelled MRI data, with generalization to out-of-sample images, such as those taken from different scanners, often quite poor (Wegmayr et al. (2018), Korolev et al. (2017)).

ii) Standard convolutional architectures cannot flexibly incorporate non-imaging data (Liu, (2019)). This is particularly problematic for neuroimaging tasks, where clinical and demographic information constitute useful sources of supplementary information that can help guide a classification decision.

iii) Deep neural networks often face the criticism that they are ‘black-box’, producing highly accurate but inscrutable predictions (Zeiler and Fergus, (2014), Gilpin et al. (2019)). Although this is not always an issue in certain accuracy-at-any-cost applications, algorithmic transparency is particularly critical when employed in clinical settings to engender trust in the diagnostic decision.

iv) GPU memory limitations restrict the possible size of mini-batches used in stochastic gradient-based optimisers to a small fraction of that possible with 2D images, resulting in high-variance gradient estimates and unstable learning. This makes networks increasingly sensitive to optimization hyper-parameters such as the learning rate and its scheduling which, if not set correctly, can result in the algorithm failing to converge. 

In order to overcome these limitations we introduce NEURO-DRAM, a 3D recurrent visual attention model tailored for neuroimaging classification. NEURO-DRAM comprises an agent which, trained by reinforcement learning, learns to attend to highly informative volumes within an image, while avoiding those providing little discriminative value. Because the agent doesn’t process the entire image in a single step, both the number of trainable parameters and the amount of computation is much lower than with fully convolutional architectures, leading to faster convergence and better generalization. Non-imaging information such as clinical and demographic data can also be incorporated by the model in a natural manner, providing hints on where to look without being directly used for the final classification decision. Finally, by visualising the agent’s trajectory, the most discriminative anatomical regions for a given classification task can be obtained, providing a clear insight into algorithm's decision. 

When applied to the task of Alzheimer's disease prediction, NEURO-DRAM achieves state-of-the-art classification accuracy, significantly outperforming a baseline convolutional neural network. NEURO-DRAM also generalizes to out-of-sample data, exhibiting no performance reduction with images taken from an external dataset. When further applied to the task of predicting which patients with mild cognitive impairment will be diagnosed with Alzheimer's disease within two years, the agent again achieves state-of-the-art accuracy with no additional training. Encouragingly, the agent learns, without explicit instruction, a search policy visiting standardized radiological hallmarks of Alzheimer’s disease, suggesting a route to automated biomarker discovery for more poorly understood disorders.

\section{Previous Work}

Several attempts have been made to address the issues outlined in the previous section within the context of neuroimaging. Spasov et al. (2019) tackled the problem of overfitting by introducing a parameter-efficient network incorporating separable convolutions and multi-task learning. Large strides and frequent down-sampling operations were also employed to help reduce the number of trainable parameters for their network, however this comes at the expense of spatial resolution, making this approach unsuitable for detecting visually subtle disorders. Data augmentation is another technique for reducing the extent of overfitting, with several authors applying transformations such as deformation, rotation, hemisphere flipping, scaling, and cropping to images before training (B\"{o}hle et al. (2019), Esmaeilzadeh et al. (2018), Wegmayr et al. (2018), Basaia et al. (2019)). However, with registration to a reference template a common preprocessing step, translations and rotations are not well motivated operations for neuroimaging data. Furthermore, deformations and cropping are unsuitable for neuroimaging tasks involving patients with severe neuronal atrophy or post-operative cavities, limiting the applicability of this approach. 

B\"{o}hle et al. (2019) addressed the interpretability issue by using layer-wise relevance propagation (LRP) to help visualise the decision of a convolutional Alzheimer's disease classifier. Esmaeilzadeh et al. (2018) took a different approach, performing an occlusion analysis to elucidate their classifier’s prediction. However, LRP performs a non-unique back-propagation from model parameters to image voxels (Bach et al. (2015)), while occlusion analysis is confounded by the issues highlighted above for cropping.

Several attempts have been made to incorporate clinical and demographic information into neuroimaging classifiers (Spasov et al. 2019, Esmaeilzadeh et al. 2018), however in these cases this was achieved by concatenating the information with an image representation learned by a separate convolutional network. When the non-imaging information is sufficiently informative, this approach offers the model a ‘shortcut’, whereby it can ignore the imaging information and instead base its decision solely on the clinical data. Behaviour of this kind is both undesirable, and clearly inflexible to variable, missing or incomplete clinical information in out-of-sample datasets.

The high computational cost of deep convolutional networks has  motivated the introduction of visual attention mechanisms for computer vision tasks. Mnih et al. (2014) introduced a recurrent attention model (RAM) which was shown to learn successful policies for cluttered and translated digit classification tasks, significantly outperforming a baseline convolutional model. Ba et al. (2015) extended RAM to recognize multiple objects in an image by including a context network to provide hints about informative regions of the image. This so-called deep recurrent attention model, or DRAM, outperformed state-of-the-art convolutional networks in a number of transcription tasks, requiring fewer parameters and less computation.

Our work seeks to address these issues simultaneously, building on the work of Mnih et al. (2014) and Ba et al. (2015) by tailoring the deep recurrent visual attention model for use with volumetric images. Like Spasov et al. (2019) we incorporate clinical, demographic, and other non-imaging data into our classifier, but in a more natural manner that doesn't allow any shortcut solutions, and is robust to variable, missing or incomplete information. Finally, we take the view that the most convincing approach to model explainability is to directly visualise the regions which proved most informative, rather than attempt to analyse a model's parameters. As such, we interpret our model’s trajectory through an image as a natural form of explainability, and suggest a route to automated biomarker discovery.  

\section{The 3D recurrent visual attention model}
NEURO-DRAM is built around a two-layer recurrent neural network (Fig. 1). At each time step $t$ the agent receives a small fraction of the entire image, centred around position $l_t$, and uses this information to decide which location to attend to at the next step $t+1$. After a fixed number of such steps the agent makes a classification decision, receiving at this point a scalar reward determined by its classification accuracy. The goal of the agent is to maximize these rewards along its trajectory, and by doing so it learns to attend to the most informative regions of the image for the task at hand. Because the model is comprised of several distinct sub-networks, we discuss each of these in turn below.
\subsection*{Glimpse network}
The glimpse network is a non-linear mapping that takes as input a small image volume $x_t$, hereafter a glimpse, as well as the corresponding location coordinate $l_t$, and outputs a vector $g_t$. The goal of the glimpse network is to learn a useful representation for $g_t$, summarising what it has seen and where it has seen it. This is performed in two steps: i) the glimpse $x_t$ is processed by a 3D convolutional neural network, comprising blocks of convolutional layers with batch normalisation and max pooling, producing a ‘what’ representation $g_{x,t}$.  ii) in parallel, the location coordinate $l_t$ is mapped to a ‘where’ vector representation $g_{l,t}$, by a single-layer fully connected neural network. Following Ba et al. (2015) and Larochelle \& Hinton (2010), $g_t$ is produced by element wise multiplication of the ‘what’ and ‘where’ representations:

\begin{equation}
g_t = g_{x,t}\odot{}g_{l,t}
\end{equation}

\begin{figure}[ht]
\centering
\includegraphics[height=8cm]{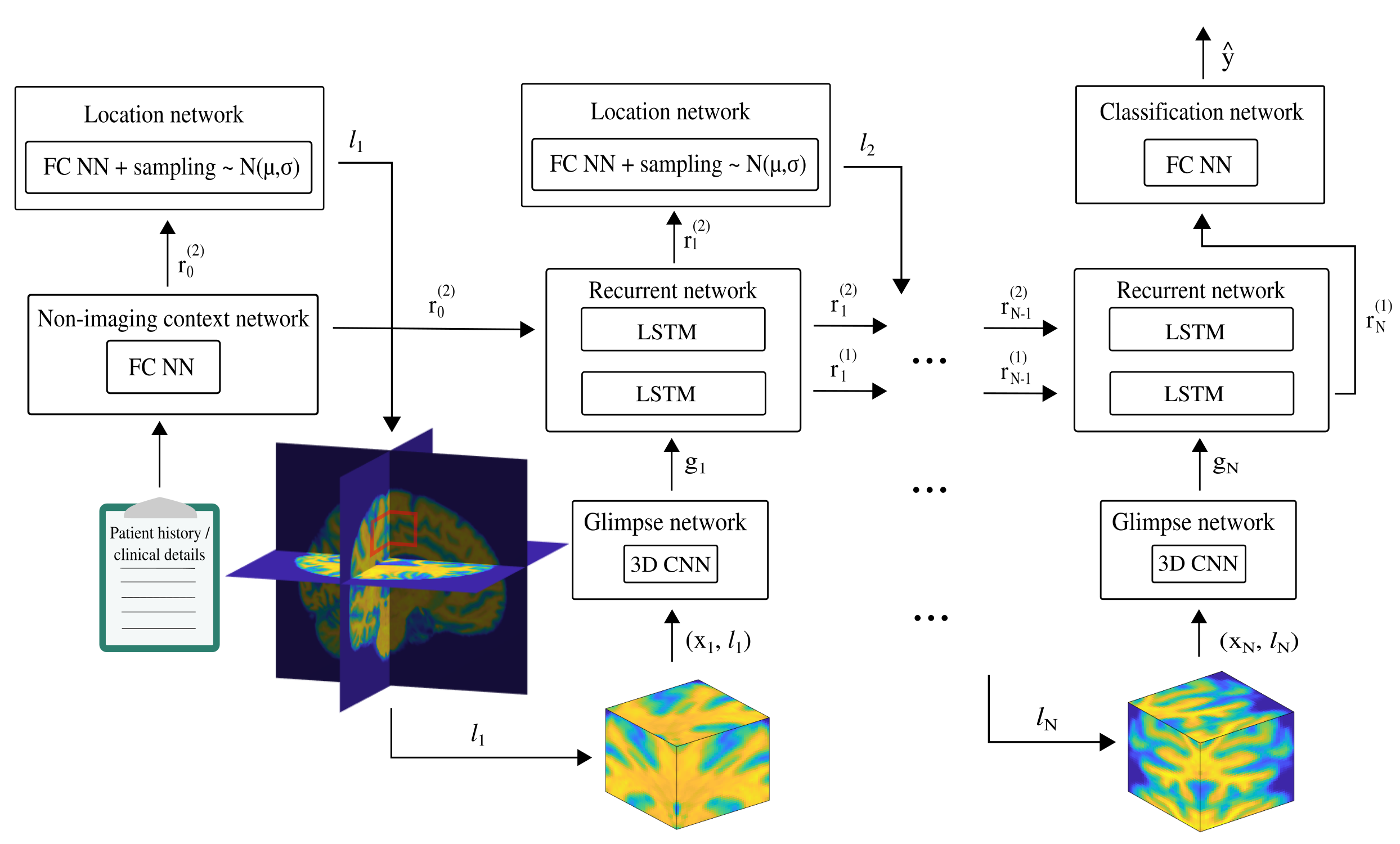}%
\caption{ Schematic representation of NEURO-DRAM. At each time step the agent selects a location to attend to and receives a small ‘glimpse’ centred at this point, outputting a classification decision after a fixed number of such steps. Trained by reinforcement learning, the agent learns to attend to informative regions for a given task, avoiding those providing little discriminative value.}
\end{figure}

\subsection*{Recurrent network}

A recurrent neural network is used to build up the agent’s internal representation summarizing information extracted from past observations. This representation is used by both the location network, which uses it to determine the next location to attend to $l_{t+1}$, and by the classification network which uses it to make a diagnostic decision at the final step. In the present work the recurrent network consists of two stacked LSTM units (Hochreiter \& Schmidhuber, 1997), employed because of their ability to learn long-range temporal dependencies. The lower LSTM cell takes as input the glimpse representation $g_t$, producing output $r_t^{1}$ which is ultimately used for classification, while the second cell takes as  input $r_t^{1}$, producing $r_t^{2}$ which is used by the location network (Fig. 1). 
This division ensures that the context network cannot directly influence the classification decision, as we discuss shortly. 

\subsection*{Location network}

The location network consists of a single-layer fully connected network which maps $r_t^{2}$ to a 3D vector $\mu_t$ in the range [-1,1] which defines the mean of an isotropic 3D normal distribution. $l_{t+1}$ is then produced by sampling from this parameterized Gaussian, with the variance a fixed hyper-parameter which controls the exploration/exploitation trade-off ubiquitous in reinforcement learning (S. Thrun, 1992). This stochasticity in determining the next location helps improve the generalisation performance of the model, as we describe later.
\subsection*{Non-imaging context network}
A context network was first introduced by Ba et al. (2015) to provide the initial state for their recurrent network. In that work, the context network was a shallow 2D convolutional network, taking as input a down-sampled (i.e., low resolution) view of the entire image, and was tasked with providing both the initial glimpse location $l_0$ as well as hints about the locations of potentially informative regions in the image. Inspired by this we introduce a non-imaging context network consisting of a single-layer fully connected network which, in the present work, produces the initial recurrent network state $r_0^{2}$ from available clinical and demographic information. This contextual information can be used by the agent to help condition its trajectory, but cannot be used by the classification network to inform the final prediction, as we discus below. The input to the non-imaging context network is not limited to demographic and clinical information, however, and could instead include, for example, radiological reports, patient clinical history, genetic information and DICOM meta-data. 

\subsection*{Classification network}

The classification network consists of a single-layer fully connected neural network with a sigmoid activation function and is used for binary classification, taking as input the final lower LSTM state vector $r_N^{1}$. As observed by Ba et al. (2015), we found that if the classification network was allowed to use the representation produced by the context network (in our case clinical and demographic details), the network could take a shortcut and learn to base its decision entirely on this information. In order to prevent this undesirable behaviour, the context and classification networks are connected to different layers of the recurrent network. As such, gradients cannot flow between the classification and context network, thereby preventing classification optimization on the basis of non-imaging data from occurring. Instead, the non-imaging information only indirectly influences the classification decision by directing where the agent moves, and therefore which image volumes it utilises. It is plausible that this is closer to the manner in which such supplementary information is incorporated by a clinical radiologist.
\subsection*{Reward}

At each time step, in addition to receiving a new glimpse $x_t$ the agent receives a scalar reward $r_t$, and its goal is to maximise the sum of these rewards over the course of its trajectory. In the case of neuroimaging classification, we set $r_t$ equal to 1 for all $t$ if after the final step the classification (i.e., Alzheimer’s disease versus healthy, brain tumour versus no brain tumour etc.) is correct, and 0 otherwise. 

At a high level, NEURO-DRAM can be considered to consist of two collaborative sub-units, one comprising the context network, glimpse network, and location network, which is tasked with providing informative visual observations to a second unit, the classification network, which utilizes these observations to make a diagnostic decision. The model is therefore a visual-based neuroimaging classifier, with the classification decision based on small but informative MRI volumes, and not clinical or demographic information.

\section{Training}
The goal of training is to learn model parameters which maximise the expected reward that the agent receives by traversing through an image. However, the sampling procedure performed by the location network introduces discontinuities into the model, precluding the use of standard gradient backpropagation techniques for parameter optimization. Instead, we employ a hybrid approach, training the classification network and glimpse networks by standard maximum-likelihood supervised learning, and the location and context networks by reinforcement learning. 

The classification and glimpse network parameters are optimized by minimizing the binary cross entropy loss between the empirical and predicted label distributions, 
\begin{equation}
-\sum\limits_{i=1}^N (y_i\log(p_i)+(1-y_i)\log(1-p_i)),
\end{equation}
where $y_i \in \{0,1\}$ is the binary label for example $i$, and $p_i$ is the probability, as predicted by the classifier network, that the $i$'th example belongs to the positive class. The gradient of this loss can be back-propagated through these two networks using standard gradient based methods like stochastic gradient descent, or using more advanced optimization procedures such as ADAM (Kingma et al. 2014) as we do in the experiments below. 

Conversely, the parameters $\theta$ for the location network and context network are trained by maximising the expectation of the total reward $J$, which in turn depends on the classification accuracy,

\begin{equation}
J(\theta) = \mathbb{E}_{\tau\sim{}\pi_{\theta}(\tau)}\left[\sum\limits_{t=1}^Tr_t\right] = \mathbb{E}_{\tau\sim{}\pi_{\theta}(\tau)}\left[R\right],
\end{equation}

with $\tau$ denoting the sequence of states and actions during the agent's trajectory, and $\pi_{\theta}(\tau)$ the policy-dependent distribution over these trajectories. Following Mnih et al. (2014) we employ the REINFORCE algorithm (Williams, 1992), one of a family of policy gradient methods from the reinforcement learning literature, which approximates the gradient of Eq. 3 by drawing $M$ trajectory samples
\begin{align}
\nabla{}J &= \sum\limits_{t=1}^T \mathbb{E}_{\tau\sim{}\pi_{\theta}(\tau)}\left[\nabla_\theta \log \pi_{\theta}(l_t\vert{}s_t)R\right],\\
&\approx \frac{1}{M}\sum\limits_{i=1}^M\sum\limits_{t=1}^T\nabla\log \pi_\theta(l_t^i|s_t^i)R^i,
\end{align}
with $\pi$ the agent's policy mapping states to actions (in our case the location network), $s_t$ the internal representation of the recurrent network at time $t$, and $i$ indexing sample number. Although seemingly complex, Eq. 4 has a simple and pleasing interpretation: at each stage in the learning procedure trajectories through a volumetric image are sampled from the current policy, with the parameters adjusted such that actions (i.e location decisions) that led to a large reward are made more likely, and those which didn't are made less likely. In this way the agent learns to attend to the most informative regions of the image, while avoiding those that provide little discriminative value.
\section{Data and experimental details}
To evaluate the effectiveness of our approach, we sought a disorder for which biomarkers have been independently established and deep learning-based approaches for its diagnosis appear in the literature. As such, we elected to evaluate NEURO-DRAM on the task of predicting Alzheimer’s disease from structural magnetic resonance images, using data from the Alzheimer’s Disease Neuroimaging Initiative (ADNI) database. Although many publications have utilised this dataset for deep learning-based classification, an objective comparison between these results and ours is challenging since each of these separately employed i) different (and often unspecified) subsets of patients, ii) different preprocessing pipelines, iii) different model inputs, and iv) different evaluation and cross-validation procedures. We combat this issue of methodological heterogeneity, highlighted previously by Samper-Gonz\'{a}lez et al. (2017), by instead fixing the architecture of our glimpse network to that of a state-of-the-art convolutional model (B\"{o}hle et al, 2019), training and evaluating both models using the same data and learning parameters (see Appendix 1 for a discussion about baseline selection). This comparative approach, which is common in the machine learning literature, allows for a fair comparison between the two models and avoids the well known phenomenon of over-fitting to increasingly stale test sets through excessive hyper-parameter tuning (Recht et al. 2019). We therefore restrict the scope of the present work to demonstrating the applicability of the approach for an archetypical neuroimaging task, opening it up to other researchers to experiment with NEURO-DRAM in other contexts, and to optimize hyper-parameter settings to maximise classification performance. A NEURO-DRAM pytorch class is made available for this purpose at \url{https://github.com/neurodram/}.

\subsection{Data and preprocessing}

The data used in this article was obtained from the Alzheimer's Disease Neuroimaging Initiative (ADNI), which can be accessed at \url{http://adni.loni.usc.edu}. For our experiments we included subjects from the ADNI-1 and ADNI-2 cohorts who were labelled as either ‘Alzheimer’s disease’ or ‘healthy control’. All scans were acquired using 1.5 Tesla scanners, and had undergone gradient non-linearity, intensity homogeneity, and phantom-based distortion correction. For all patients we downloaded the Magnetization Prepared Rapid Gradient-Echo (MPRAGE) T1-weighted images, along with each patient’s demographic data (gender, age, ethnic and racial categories, and education), cognitive performance measures (CDRSB, ADAS11, ADAS13, RAVLT), and APOe4 expression indicator. In total, 322 patients were included (162 AD, 160 HC), for a total of 1012 images. These were split into a training set (109 AD patients, 107 HCs; 797 images in total), a validation set (18 AD patients, 18 HCs; 100 images in total) and a test set (35 AD patients, 35 HCs; 70 images in total). Following B\"{o}hle et al. (2019), we performed the data split at the level of participants to ensure independence between training, validation, and test sets, with both models trained and evaluated using these fixed data splits. Once downloaded, we perform brain extraction using DeepBrain, an open source CNN-based skull stripping tool available at \url{https://github.com/iitzco/deepbrain}, as well as non-linear registration to the reference Montreal Neurological Institute (MNI) 1 mm T1 brain template using ANTsPy, an open-source python wrapper for Advanced Normalization Tools (Avants, 2009), available at \url{https://github.com/ANTsX/ANTsPy}.

\subsection{Network architectures and training details}

The baseline model is a convolutional neural network consisting of four convolutional blocks, followed by two fully connected layers (B\"{o}hle et al. (2019)), taking as input the entire image volume. Each block contains f filters (f = 8, 16, 32,64) of size 3 x 3 x 3, followed by batch normalization (Ioffe and Szegedy, 2015) and max pooling (window size 4 x 4 x 4), while the fully connected layers contain 128 and 2 units, respectively. A softmax activation function provides the probabilities for the two classes, and the whole network is trained using the ADAM optimizer, with an initial learning rate of $10^{-4}$ and exponential decay scheduling. Dropout is applied to the fully connected layers (p = 40\%) during training, and $\textrm{L}_2$ regularization is employed with a weight decay constant $\lambda=10^{-4}$.

The architecture of our glimpse network was fixed to that of the baseline described above. All hidden layers in the location, recurrent, classification, and context networks had a  dimensionality of 512, with no experimentation of these settings performed. The number of glimpses the agent receives was set to 6, with each glimpse of size 40 x 40 x 40 ($\sim 1.6\%$ of the entire image volume). No regularization in the form of weight decay or drop-out was included. All other learning parameters such as the optimization algorithm, learning rate, and scheduling were fixed to that of the baseline. 

Both models were trained on the same training dataset using an NVIDIA RTX 2080Ti graphics card, with no data augmentation performed. Early stopping was employed, with the best performing checkpoint on the validation set for each model used to assess the final classification performance on the same independent test set. 

\section{Results}

NEURO-DRAM significantly outperforms the baseline model for Alzheimer’s disease prediction, demonstrating a classification accuracy of 98.5\% on the ADNI test set (Table 1). This is achieved despite the fact that the agent sees less than 10\% of each image, and without optimization of architectural and learning hyper-parameters. By construction, the agent only uses the imaging information to make its decision. We confirmed this behaviour by blanking the images and providing only the clinical context information; in this case the algorithm performs no better than chance ($54\%$), as expected. Furthermore, we verified that backpropagated parameter gradients cannot flow from the classification network to the non-imaging context network, confirming that classification optimization on the basis of non-imaging information isn't occurring. 

\begin{table}[h]
    \begin{center}
        \begin{tabular}{| l | l | l |}
        \hline
        Model & NEURO-DRAM & CNN baseline  \\ \hline
        Balanced accuracy & 98.5\% & 87.5\%  \\ 
        \hline
        Specificity & 100\% & 82\% \\
        \hline
        Sensitivity & 97.1\% & 86\% \\ \hline
        Maximum mini-batch size & 60 & 6 \\ \hline
        Training time (per epoch) & 45 seconds & 241 seconds\\
        \hline
        \end{tabular}
        \vspace{1em}
    \caption{Classification performance of NEURO-DRAM and a fully convolutional baseline (B\"{o}hle (2019)).} 
    \end{center}
\end{table}

NEURO-DRAM also out-performs all previously reported deep-learning-based models, including those incorporating both structural MRI and PET data (Jo et al. (2019) (Table 2). 

\begin{table}[h]
    \begin{center}
        \begin{tabular}{| l | l | l | l |}
        \hline
        Reference & Modality & Model & AD vs. HC accuracy \\ \hline
        Current study & MRI & NEURO-DRAM & 98.5\%  \\ 
        \hline
        Cheng et al. (2018) & MRI & 3D CNN & 87.15\%  \\ \hline
        Wegmayr et al. (2017) & MRI &3D CNN &  86\%  \\ \hline
        Korolev et al. (2017) & MRI &3D CNN &  80\%  \\ \hline
        Suk et al. (2015) & MRI, PET & DBM & 95.35\% \\ \hline
        Li et al (2014) & MRI, PET & SAE & 92.8\%   \\ \hline
        Liu et al (2015) & MRI, PET & SAE & 91.4\%   \\ \hline
        Vu et al. (2017) & MRI, PET & SAE + 3D CNN & 91.14\%\\ \hline
        Cheng and Liu (2017) & MRI, PET & 3D CNN + 2D CNN & 89.6\% \\ \hline
        Lu et al. (2018) & MRI, PET & DNN + NN & 84.6\%  \\ \hline
        \end{tabular}
        \vspace{1em}
    \caption{Comparison between NEURO-DRAM and previously reported models.} 
    \end{center}
\end{table}\vspace{1em}

To test the generalization performance of NEURO-DRAM on truly independent, out-of-sample data, we applied it to subjects from the Open Access Series of Imaging Studies (OASIS) database (all participants from the OASIS-1 cohort were included, consisting of 100 AD and 130 HC, one image per participant, available at \url{http://www.oasis-brains.org/}).
Because less additional non-imaging data is available for images from this database (it is missing, for example, genetic and ethnic information, as well as cognitive performance measures like ADAS and RAVLT scores), this task also provides a test of our model's ability to handle variable, missing or incomplete clinical and demographic information. Instead of retraining the agent to incorporate only the non-imaging data common to both databases, for each OASIS test set participant we set the missing data to be equal to that of the most similar ADNI participant, as quantified by the cosine similarity across the remaining fields in common (age, gender, cognitive score (CDR), education). This allows the agent to draw on previous experience when conditioning it's trajectory, but forbids it from using this imputed data for the classification decision. It is plausible that this is similar to the manner in which a radiologist might deal with missing or incomplete non-imaging information.

NEURO-DRAM generalizes well to the independent OASIS dataset, with no reduction in classification accuracy when compared to the ADNI test set. This is in contrast to the fully convolutional baseline, which performs considerably worse  (Table 3). To the best of our knowledge, this is the highest reported generalization accuracy using an external, out-of-sample test set. 
\begin{table}[h]
    \begin{center}
        \begin{tabular}{| l | l | l |}
        \hline
        Model & NEURO-DRAM & CNN baseline \\ \hline
        Accuracy & 99.8\%  & 71.3\%\\ 
        \hline
        Specificity & 100\% & 85\%  \\ \hline
        Sensitivity & 99\% & 50\%  \\ 
        \hline
        \end{tabular}
        \vspace{1em}
    \caption{Generalization performance to OASIS test set} 
    \end{center}
\end{table}

Unlike convolutional networks, NEURO-DRAM  offers a clear insight into its classification decision. By visualising the agent’s trajectory the most discriminative anatomical regions for Alzheimer’s disease prediction can be obtained, providing a natural form of evidence for the algorithm's decision. Interestingly, the agent learns to first attend to the medial temporal lobe, before moving posteriorly to the temporoparietal cortex (Fig. 2 and Fig. 3). By following this trajectory, the agent receives glimpses of the hippocampus, parahippocampal gyrus, lateral ventricles and parietal cortex, volume loss in which regions has been correlated with Alzheimer’s disease in many studies (Kesslak et al., 1991, Smith and Jobust, 1996, Jack et al., 1997, Nagy et al., 1999, Apostolova et al., 2012, Van Hoesen et al., 2000, Jacobs et al., 2012). 
\begin{figure}[ht]
\centering
\includegraphics[height=11cm]{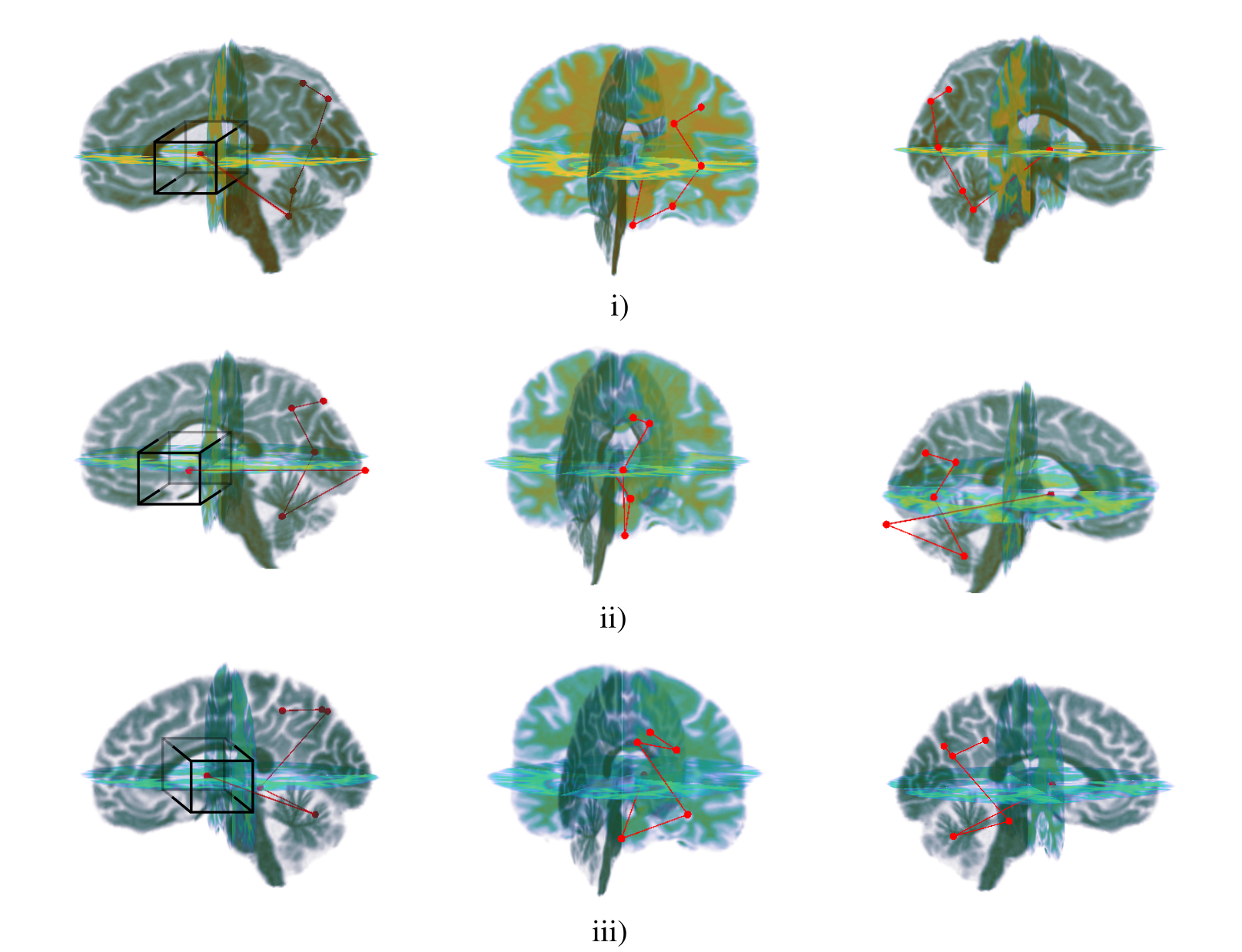}%
\caption{Trajectory taken by the agent for three representative participants from the ADNI test set (one per row). A bounding box around the first location attended to is included to indicate the approximate size of the glimpse that the agent receives; this is the same for all subsequent locations  By taking this path, the agent receives glimpses of the hippocampus, parahippocampal gyrus, lateral ventricles and parietal cortex, regions known to be affected by Alzheimer’s disease.
}
\end{figure}

\begin{figure}[ht]
\centering
\includegraphics[height=3.5cm]{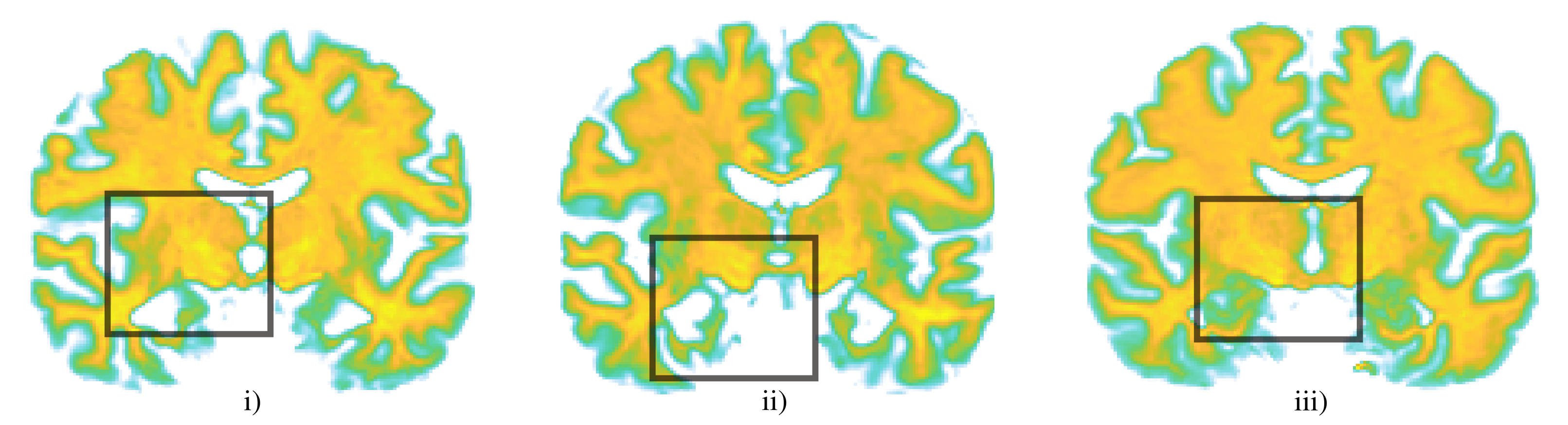}%
\caption{Coronal slice through the first glimpse taken by the agent for two Alzheimer's disease subjects, i), ii), and one healthy control, iii), from the ADNI test set. By attending to this region, the agent observes the hippocampus and parahippocampal gyrus. Atrophy in this region is a radiological biomarker of Alzheimer's disease, as can be seen in i) and ii).   
}
\end{figure}

As a final test of our agent's policy we apply it, without additional training, to the task of predicting which patients with mild cognitive impairment (MCI) will progress to Alzheimer's disease within two years of clinical assessment. Because some forms of MCI are clinical precursors of Alzheimer's disease, it is plausible that the search policy followed by our agent could also help identify those patients at risk of progression. To this end, we freeze the weights of both NEURO-DRAM and the fully convolutional baseline following AD vs. HC training, and apply these models to an independent test set comprising 36 previously unseen patients from the ADNI database, 18 of which had MCI at the time of their baseline scan and went on to develop Alzheimer’s disease within two years, and 18 of which had MCI but did not progress. Again, NEURO-DRAM significantly outperforms the baseline architecture, demonstrating a classification accuracy of 77.8\%, versus only 63\% for the fully convolutional model, suggesting that the locations attended to by the agent are informative for both early and advanced stages of the disease. To our knowledge, this is the highest classification accuracy for predicting progression of MCI from structural MRI scans only (Jo et al., (2019)), despite the fact that the agent was not explicitly trained using any MCI data.

\begin{table}[h]
    \begin{center}
        \begin{tabular}{| l | l | l | l |}
        \hline
        Reference & Modality & Model & AD vs. HC accuracy \\ \hline
        Current study & MRI & NEURO-DRAM & 77.8\%  \\ 
        \hline
        Liu et al. (2018b) & MRI & Landmark detection \& 3D CNN & 76.9\%  \\ \hline
        Basaia et al. (2019) & MRI &  3D CNN & 75\%  \\ \hline
        Spasov et al. (2010) & MRI &3D CNN &  72\%  \\ \hline
        \end{tabular}
        \vspace{1em}
    \caption{Comparison between NEURO-DRAM and previously reported models for MCI conversion prediction.} 
    \end{center}
\end{table}

\section{Discussion}
Our novel model introduced in this paper addresses a number of issues associated with using fully convolutional architectures for three-dimensional neuroimaging data. As a result, it outperforms a state-of-the-art baseline architecture at the task of Alzheimer’s disease classification, despite very limited hyper-parameter tuning. NEURO-DRAM also generalizes well to an independent dataset; an essential goal if machine-learning analysis of neuroimaging data is ever to be deployed in a clinical setting. This improved performance can be traced to the agent learning to attend to highly informative volumes within an image while avoiding those providing little discriminative value, thereby reducing the effects of overfitting.
The increased usable batch sizes and faster training of our model also affords much greater scalability to large neuroimaging datasets, with training times less than ten minutes for the dataset presented in this paper, compared with over 45 minutes for the baseline model. This is due to the smaller number of computations and more stable gradient updates performed by NEURO-DRAM. 
The model's state-of-the-art performance is also achieved without any form of data augmentation or model regularization. This is important because it requires less hyper-parameter tuning ($\textrm{L}_2$ penalty coefficient, drop-out probability, etc.) to optimize model performance, and avoids the introduction of unnatural and task-dependent data transformations such as deformations and cropping.

Perhaps most importantly, NEURO-DRAM offers a clear insight into its classification decision. While the validity of these model decisions will require further investigation, we believe NEURO-DRAM has great promise to provide sufficient interpretability to reach a level of acceptance in clinical settings. As shown in Fig. 2, the agent primarily visits the medial temporal cortex and the temporoparietal region, and by doing so receives views of the hippocampus, parahippocampal gyrus, lateral ventricles and parietal cortex, regions known to be affected by Alzheimer’s disease-related neuronal atrophy. 
In fact, clinical hallmarks for diagnosing Alzheimer’s disease are based on standardized assessments of these very regions, with the Koedam score for parietal atrophy (Koedman et al., 2011) and the MTA scale for medial temporal atrophy widely used by clinical radiologists (Scheltens et al. (1995)).
That the agent learns to attend to these known biomarkers for Alzheimer’s disease suggests that it is functioning in way that is analogous to how a radiologist may view images when considering a diagnosis of Alzheimer's disease. Furthermore, because the neuroanatomical regions utilised by the model can be readily visualised, when applied to less well-characterised psychiatric and neurological disorders the trajectory taken by the agent can be considered as representing feature/location importance, offering a novel route to automated biomarker discovery. 

Several straight-forward extensions to the existing model are also possible. A key feature of NEURO-DRAM is the ability to include non-imaging meta-data. While here we included only numerical and categorical clinical and demographic information, clinical history in the form of structured radiological reports, or even free text, could be incorporated by replacing the fully-connected context network with, for instance, a language embedding module such as word2vec (Mikolov et al 2013) or doc2vec (Le and Mikolov, 2014). NEURO-DRAM could also be employed recursively for multi-modal analysis. In this case, instead of providing a classification decision at the final step, the hidden state of the upper LSTM unit would become the initial context vector for the next modality (e.g., a PET scan, a different MRI sequence, or even a lumbar puncture for CSF biomarker analysis), guiding the search by keeping in mind what it has seen in the structural MR image. Finally, the classification network can also be modified to output a decision as soon as it is sufficiently confident of its prediction by incorporating a negative reward at each time step. In this way, the agent would attend to additional regions (or other imaging modalities) only if this will considerably improve the certainty of it's decision. This feature has potentially important implications for clinical practice, as the cost and invasiveness of each modality can be considered. This way, unnecessary clinical investigations can be reduced and only as much data as is required for a given clinical decision is acquired, reducing costs and patient discomfort.

\section{Conclusion}
Here we present NEURO-DRAM, a 3D recurrent visual attention model tailored for neuroimaging data. The results in the analysis of Alzheimer's disease and MCI suggest that this reinforcement learning approach has considerable promise for improving model performance, for combining  neuroimaging and meta-data and for providing intuitive interpretability of how the model reaches decisions.

\subsubsection*{Acknowledgments}
We would like to thank Pritesh Mehta for many helpful comments and discussions, and Jeremy Lynch for providing the computer used for training the model. This work was supported by the Wellcome/EPSRC Centre for Medical Engineering (WT 203148/Z/16/Z). \\
Data collection and sharing for this project was funded by the Alzheimer's Disease
Neuroimaging Initiative (ADNI) (National Institutes of Health Grant U01 AG024904) and
DOD ADNI (Department of Defense award number W81XWH-12-2-0012). ADNI is funded
by the National Institute on Aging, the National Institute of Biomedical Imaging and
Bioengineering, and through generous contributions from the following AbbVie, Alzheimer's
Association, Alzheimer's Drug Discovery Foundation, Araclon Biotech, BioClinica,
Biogen,  Alzheimer Immunotherapy Research and Development, LLC., Johnson and Johnson
Pharmaceutical Research and Development LLC., Lumosity, Lundbeck, Merck and Co., Inc.,
Meso Scale Diagnostics, LLC., NeuroRx Research, Neurotrack Technologies, Novartis
Pharmaceuticals Corporation, Pfizer Inc., Piramal Imaging, Servier, Takeda Pharmaceutical
Company, and Transition Therapeutics. The Canadian Institutes of Health Research is
providing funds to support ADNI clinical sites in Canada. Private sector contributions are
facilitated by the Foundation for the National Institutes of Health (\url{www.fnih.org}). The grantee
organization is the Northern California Institute for Research and Education, and the study is
coordinated by the Alzheimer's Therapeutic Research Institute at the University of Southern
California. ADNI data are disseminated by the Laboratory for Neuro Imaging at the
University of Southern California. 

\section*{References}

\medskip

\small

[1] S. Spasov, L. Passam, A. Duggento, and P. Liò. A parameter-efficient deep learning approach to predict conversion from mild cognitive impairment to Alzheimer’s disease. In \emph{Neuroimage}, 2019.

[2] V. Wegmayr, S. Aitharaju, and J. Buhmann. Classification of brain MRI with big data and deep convolutional neural networks. In \emph{Proceedings of SPIE Medical Imaging}, 2018. 

[3] S. Korolev, A. Safiullin, M. Belyaev, and Y. Dodonova. Residual and plain convolutional neural networks for 3D brain MRI classification. In \emph{ISBI}, 2017. 

[4] D. Cheng, M. Liu, J. Fu, and Y. Wang. Classification of MR brain images by combination of multi-CNNs for AD diagnosis. In \emph{Ninth International Conference on Digital Image Processing}, 2017.

[5] M. B\"{o}hle, F. Eitel, M. Weygandt, and K. Ritter. Layer-wise relevance propagation for explaining deep neural network decisions in MRI-based Alzheimer’s Disease Classification. In \emph{Frontiers in Ageing Neuroscience}, 2019

[6] S. Vieira, W. Pinaya and A. Mechelli. Using deep learning to investigate the correlates of psychiatric and neurological disorders: methods and applications. In \emph{Neuroscience and Biobehavioral Reviews}, 2017.

[7] X. Liu, L. Faes, A. Kale, S. Wagner, D. Fu, A. Bruynseels, T. Mahendiran, G. Moraes, M. Shamdas, C. Kern, J. Ledsam, M. Schmid, K. Balaskas, E. Topol, L. Bachmann, P. Keane, A. Denniston.A comparison of deep learning performance against health-care professionals in detecting diseases from medical imaging: a systematic review and meta-analysis. In \emph{The Lancet}, 2019. 

[8] M. Zeiler and R. Fergus. Visualizing and understanding convolutional networks. In \emph{European Conference on Computer Vision}, 2014.

[9] L. Gilpin, D. Bau, B. Yuan, A. Bajwa, M. Specter, and L. Kagal. Explaining explanations: an overview of interpretability of machine learning.  \emph{arXiv preprint arXiv:1806.00069}, 2014.

[10] S. Esmaeilzadeh, D. Belivanis, K. Pohl, and E. Adeli. End-to-end Alzheimer’s disease diagnosis and biomarker identification. In \emph{International Workshop on Machine Learning in Medical Imaging},  2018. 

[11] S. Basaia, F. Agosta, L. Wagner, E. Canu, G. Magnani, R. Santangelo, M. Filippi. Automated classification of Alzheimer's disease and mild cognitive impairment using a single MRI and deep neural networks. In \emph{Neuroimage:Clinical}, 2019. 

[12] S. Bach  , A. Binder , G. Montavon, F. Klauschen, K. M\"{u}ller. On Pixel-wise explanations for non-linear classifier decisions by layer-wise relevance propagation. In \emph{PLOS One}, 2015.

[13] V. Mnih, N. Heess, A. Graves, and K. Kavukcuoglu. Recurrent models of visual attention. In \emph{Advances in Neural Information Processing Systems}, 2014.

[14] J. Ba,  V. Mnih,  and K. Kavukcuoglu. Multiple object recognition with visual attention. In \emph{ICLR}, 2015

[15] H. Larochelle and G. Hinton. Learning to combine foveal glimpses with a third-order Boltzmann machine. In \emph{Advances in Neural Information Processing Systems}, 2010.

[16] S. Hochreiter, and J. Schmidhuber. Long short-term memory.  In \emph{Neural computation}, 1997.

[17] S. Thrun. Efficient exploration in reinforcement learning. Technical report. 1992.

[18] D. Kingma and J. Ba. Adam: a method for stochastic optimization. In \emph{ICLR}, 2015.

[19] R. Williams. Simple statistical gradient-following algorithms for connectionist reinforcement learning. In \emph{Machine Learning}, 1992.

[20] J. Samper-González, N. Burgos, S. Fontanella, H. Bertin, M. Habert, S. Durrleman, T. Evgeniou, and O. Colliot. Yet another ADNI machine learning paper? paving the way towards fully-reproducible research on classification of Alzheimer’s disease. \emph{arXiv preprint arXiv:1709.07267}, 2017.  

[21] B. Recht, R. Roelofs, L.Schmidt, and V. Shankar. Do ImageNet classifiers generalize to ImageNet? 
\emph{arXiv preprint arXiv:1902.10811}, 2019.

[22] B. Avants, N. Tustison and G. Song. Advanced Normalization Tools, ANTS 1.0. Sourceforge, 2009.

[23] S. Ioffe, and C. Szegedy. Batch normalization: accelerating deep network training by reducing internal covariate shift. \emph{arXiv preprint arXiv:1502.03167}, 2015.

[24] T. Jo, K. Nho, and A. Saykin. Deep learning in Alzheimer’s disease: diagnostic classification and prognostic prediction using neuroimaging data. In \emph{Frontiers in Ageing Neuroscience}, 2019.  

[25] I. Suk, S. Lee and D. Shen. Hierarchical feature representation and multimodal fusion with deep learning for AD/MCI diagnosis. In \emph{NeuroImage}, 2014.

[26] R. Li, W. Zhang, I. Suk, L. Wang, J. Li, D. Shen et al. Deep learning based imaging data completion for improved brain disease diagnosis. In \emph{International Conference on Medical Image Computing and Computer-Assisted Intervention}, 2014.

[27] S. Liu, W. Cai, H. Che, S. Pujol, R. Kikinis et al. Multimodal neuroimaging feature learning for multiclass diagnosis of Alzheimer's Disease. In \emph{IEEE Transactions of Biomedical Engineering}, 2015.

[28] T. Vu, H. Yang, V. Nguyen, A. Oh, and M. Kim. Multimodal learning using convolution neural network and Sparse Autoencoder. In \emph{IEEE International Conference on Big Data and Smart Computing}, 2017.

[29] D. Lu, K. Popuri, G. Ding, R. Balachandar, and M. Beg. Multimodal and multiscale deep neural networks for the early diagnosis of Alzheimer's disease using structural MR and FDG-PET images. In \emph{Scientific Reports}, 2018.

[30] M. Buchsbaum, J. Kesslak, G. Lynch, H. Chui, J. Wu, N. Sicotte, E. Hazlett, E. Teng, C. Cotman. Temporal and hippocampal metabolic rate during an olfactory memory task assessed by positron emission tomography in patients with dementia of the Alzheimer type and controls. Preliminary studies.In \emph{Archives of General Psychiatry}, 1991. 

[31] A. Smith and K. Jobust. Use of structural imaging to study the progression of Alzheimer's disease. In \emph{British Medical Bulletin}, 1996.

[32] C. Jack, R. Petersen, P. O’Brien, and E. Tangalos. MR-Based Hippocampal volumetry in the diagnosis of Alzheimer’s disease. In \emph{Neurology}, 1992.

[33] Z. Nagy, N. Hindley, H. Braak, E. Braak, D. Yilmazer-Hanke, C. Shultz, et al. The progression of Alzheimer’s disease from limbic regions to the neocortex: clinical, radiological, and pathological relationships. In \emph{Dementia and Geriatric Cognitive Disorders}, 1999. 

[34] L. Apostolova, A. Green, S. Babakchanian, K. Hwang, Y. Chou, A. Toga, et al. Hippocampal atrophy and ventricular enlargement in normal ageing, mild cognitive impairment, and Alzheimer’s disease. In \emph{Alzherimer’s Disease and Associated Disorders}, 2012. 

[35] G. Van Hoesen, J. Parvizi, and C. Chu. Orbitofrontal cortex pathology in Alzheimer's disease. In \emph{Cerebral Cortex}, 2000. 

[36] H. Jacobs, M. Van Boxtel, J. Jolles, F. Verhey, and H. Uylings. Parietal cortex matters in Alzheimer's disease: an overview of structural, functional and metabolic findings. In \emph{Neuroscience and Biobehavioral Reviews}, 2012.  

[37] E. Koedman, M. Lehmann, W. van der Flier, P. Scheltens, Y. Pijnenburg, N. Fox, F. Barkhof, and M. Wattjes. Visual assessment of posterior atrophy development of a MRI rating scale. In \emph{European Radiology}, 2011. 

[38] P. Scheltens, L. Launer, F. Barkhof et-al. Visual assessment of medial temporal lobe atrophy on magnetic resonance imaging: Interobserver reliability. In \emph {Journal of Neurology}, 1995.

[39] T. Mikolov, I. Sutskever, K. Chen, G. Corrado, J. Dean. Distributed representations of words and phrases and their compositionality. In \emph{Advances in Neural Information Processing Systems}, 2013.

[40] Q. Le and T. Mikolov. Distributed representations of sentences and documents. \emph{arXiv preprint arXiv:1405.4053}, 2014.

[41] A. Gupta, M. Ayhan, and A. Maida. Natural image bases to represent neuroimaging data. In \emph{Proceedings of the 30th International Conference on Machine Learning}, 2013.

[42] A. Payan, and G. Montana. Predicting Alzheimer’s disease: a neuroimaging study with 3D convolutional neural networks. \emph{arXiv preprint arXiv:1502.02506}, 2015. 

[43] E. Hosseini, G. Gimel’farb, and A. El-Baz. Alzheimer’s disease diagnostics by deeply supervised adaptable 3D convolutional network. \emph{arXiv preprint arXiv:1607.00556}, 2016.

\section{Appendix}

Selection of Baseline model

This model was selected on the basis of classification accuracy, taking into account methodological transparency. Although several papers have reported classification accuracies above 90\% using multiple scans per participant (Gupta et al. (2013), Payen et al. (2015), Hosseini et al. (2016), and Basaia et al. (2019)), in all cases the authors failed to explicitly state how the split into training, validation, and test sets was performed. As shown by Wagmayr et al. (2018), if the data is shuffled at the level of images rather than participants, classification accuracy dramatically increases - in their case from 86\% to over 98\%. This occurs because the model is seeing participants in the test set that it saw during training, resulting in an overly optimistic estimate of its generalization performance. Interestingly, all publications which explicitly state that the correct data split was performed achieved significantly lower classification accuracies, calling into question the performance of those that failed to do so. For this reason, the models of Gupta et al. (2013), Payen et al. (2015), Hosseini et al. (2016), and Basaia et al. (2019) was deemed ineligible for consideration as a state-of-the-art baseline. 

Of the remaining publications that are methodologically transparent, B\"{o}hle et al. (2019) reported the highest classification accuracy. As such, we selected their model as a state-of-the-art baseline for comparison with NEURO-DRAM.

\end{document}